\documentclass[11pt]{article}

\usepackage{lipsum}

\usepackage{booktabs}       
\usepackage{color}
  \usepackage{nopageno}   
\usepackage{amsmath,amssymb,amsthm}
\usepackage{amsmath,amssymb,mathtools}
\usepackage{bm,xparse}
\usepackage{amsmath}
\usepackage{amssymb}
\usepackage{verbatim}
\usepackage{subfigure}
\usepackage[final]{graphicx}
\usepackage{psfrag}
\usepackage{xcolor}
\usepackage{lipsum}
\usepackage{authblk}
\usepackage{blindtext}
\usepackage{geometry}
\usepackage{mathrsfs}
\usepackage{cite}
\usepackage{dsfont}
\usepackage{url}
\usepackage{fullpage}
\usepackage{color}
\usepackage{amsmath,amssymb,amsthm}
\usepackage{amsmath,amssymb,mathtools}
\usepackage{bm}
\usepackage{amsmath}
\usepackage{amssymb}
\usepackage{latexsym}
\usepackage{verbatim}
\usepackage{subfigure}
\usepackage[final]{graphicx}
\usepackage{psfrag}
\usepackage{xcolor}
\usepackage{lipsum}
\usepackage{authblk}
\usepackage{geometry}
\usepackage{mathrsfs}
\usepackage{bbm}
\renewcommand{\cal}{\mathcal}

\newcommand{\cC}{{\cal C}}
\newcommand{\cF}{{\cal F}}
\newcommand{\cY}{{\cal Y}}
\newcommand{\cD}{{\cal D}}

\newcommand\cH{{\mathcal H}}

\newcommand{\cM}{{\cal M}}

\newcommand{\cV}{{\mathcal V}}

\newcommand{\cX}{{\mathcal X}}

\NewDocumentCommand{\errrob}{ O{\infty} O{\epsilon} 
}{\mathsf{err}_{\textup{robust}}^{#1,#2}}

\newcommand{\bE}{\mathbb{E}}

\newcommand{\bP}{\mathbb{P}}

\newcommand{\bR}{{\mathbb R}}

\newcommand{\sD}{{\mathscr{ D}}}

\renewcommand{\leq}{\leqslant}
\renewcommand{\geq}{\geqslant}

\newcommand{\cnote}[1]{\textcolor{cyan}{[CD:~#1]}}

\newtheorem{theorem}{Theorem}[section]

\newtheorem{remark}{Remark}[section]
\newtheorem{assumption}{Assumption}[section]

\newtheorem{definition}{Definition}[section]

\usepackage{algorithm, algorithmic}

\let\Pr\relax
\DeclareMathOperator*{\Pr}{\mathbb{P}}

\DeclareMathOperator{\argmax}{argmax}

\newcommand{\bI}{{\mathbb I}}

\begin{document}
\title{Representation via Representations: \\Domain Generalization via Adversarially Learned Invariant Representations }

%

%
\author[1]{Zhun Deng
\thanks{Author names listed in alphabetical order. Corresponding authors: {\tt  zhundeng@g.harvard.edu}, {\tt dwork@seas.harvard.edu}, {\tt pragya@seas.harvard.edu}.}}
\author[2]{Frances Ding }
\author[1]{Cynthia Dwork}
\author[1]{Rachel Hong}
\author[1]{ Giovanni Parmigiani }
\author[3]{ Prasad Patil }
\author[1]{Pragya Sur}
\affil[1]{Harvard University}
\affil[2]{University of California, Berkeley}
\affil[3]{Boston University}
\date{}

\maketitle

\begin{abstract}
We investigate the power of censoring techniques, first developed for learning {\em fair representations}, to address domain generalization. 
We examine {\em adversarial} censoring techniques for learning  invariant representations from multiple "studies" (or domains), where each study is drawn according to a distribution on domains.  The mapping is used at test time to classify instances from a new domain.
In many contexts, such as medical forecasting, domain generalization from studies in populous areas (where data are plentiful), to geographically remote populations (for which no training data exist) provides fairness of a different flavor, not anticipated in previous work on algorithmic fairness.

We study an adversarial loss function for $k$ domains and precisely characterize its limiting behavior as $k$ grows, formalizing and proving the intuition, backed by experiments, that observing data from a larger number of domains helps. The limiting results are accompanied by non-asymptotic learning-theoretic bounds. Furthermore, we obtain sufficient conditions for good  worst-case prediction performance of our algorithm on previously unseen domains. Finally, we decompose our mappings into two components and provide a complete characterization of invariance in terms of this decomposition. To our knowledge, our results provide the first formal guarantees of these kinds for adversarial invariant domain generalization. 

\end{abstract}

\section{Introduction}\label{sec:introduction}
\newcommand{\etal}{{\it et al.}}
\newcommand{\RvR}{Representation via Representations}
\newcommand{\remove}[1]{}
\newcommand{\gstar}{g^*}
\newcommand{\disp}{\mathrm{dys}}
\newcommand{\mI}{\mathds{1}}


In gene expression analysis, as well as in much of high-throughput biology analyses on human populations, variation between studies can arise from the intrinsic biological heterogeneity of the populations being studied, or from technological differences in data acquisition. In turn both these types of variation can be shared across studies or not.  For example, an algorithm for predicting whether a tumor will recur, trained on data obtained from the local population via a specific data-collection and processing method at a research hospital $A$, will typically not perform equally well on data collected at a research hospital $B$, using different data-collection techniques and serving a potentially different local population.  

Of course, theoretically, $B$ can train its own algorithm.  This ``siloization'' is suboptimal for several reasons, from reduced statistical power, to wasteful allocation of research investment due to duplication of effort, or even reluctance to fund or publish such duplication.  It is preferable to combine the data sources, potentially with some smart provision for domain variation.  However, there will always be a new $C$ to which the resulting algorithm will need to be applied -- maybe the competition across town who did not collaborate at the development stage, or maybe a small, geographically isolated, population, far from any major medical research center.
The ultimate goal, in the development of models with biomedical applications, is to provide accurate predictions for fully independent samples, originating from institutions and processed by laboratories that did not generate the training datasets.  How can we transfer prediction capability to a new population?

This is a problem of {\em domain generalization}, the subject of intense study for nearly two decades \cite{blanchard2011generalizing,muandet2013domain,li2018deep,li2018domain,khosla2012undoing,xu2014exploiting,niu2015multi,niu2016exemplar,ghifary2016scatter,ghifary2015domain,volpi2018generalizing,erfani2016robust,deng2018image,shankar2018generalizing,li2017deeper,akuzawa2018domain,matsuura2019domain}.
Under the assumption that there is a common signal that provides a high quality predictor $\gstar$ for {\em all} populations, and given labeled training data from several populations, can this signal be learned even when it does not necessarily yield the best predictor for any given population? When does the presence of multiple training datasets improve the accuracy of this learning procedure?

Using tools developed for finding "fair" representations of individuals in which sensitive attributes such as sex or race have been censored~\cite{Zemel:2013wz,Louizos:2015te,Edwards:2015vn,madras2018learning}, we proceed from the following intuition: treating the domain as a sensitive attribute and training on multiple, highly diverse, populations, the learning algorithm is forced to disregard the idiosyncratic in favor of the universal, 
that is, to find a prediction rule based on a signal that is shared among all domains.


This work -- domain generalization to unseen populations -- provides a new dimension of fairness, {\em transferring the benefit of federal research dollars from preeminent bench to geographically remote bedside}, not anticipated in earlier work on learning fair representations.

\paragraph{Approach.} We model the problem through the lens of a hierarchical Bayesian approach that is extensively used in applications. Let $\cX\subseteq \bR^d$ be the covariate space and $\cY=\{0,1\}$ the outcome space.  Let $\sD$ denote a collection of probability distributions on $\mathcal{X} \times \mathcal{Y}$ and $\mu$ be a distribution supported on $\sD$. The observed data arises through a hierarchical scheme---first, domains
 $\cD_1,\cdots,\cD_k$  are sampled i.i.d.~from $\mu$,  
 and then random samples $S_i=\{x_{i,j},y_{i,j}\}_{j=1}^{n_{i}}$ are drawn from each $\cD_i$.  We seek to train a classifier on the observed samples that performs well on any distribution from $\sD$, even those from which no data have been observed. To this end, we adopt an adversarial censored learning approach. 
 Simplifying slightly, for 
a mapping $\phi$ from the input covariate space to a representation space $Z$, a discriminator $\psi_k$ that attempts to guess the source domain of $\phi^{-1}(z)$ for $z \in Z$, and a classifier $f$,
we define an empirical adversarial loss function  
that increases with misclassifications by $f$ and correct guesses by the discriminator (Equation~\ref{eq:empirical_loss}). Our approach then tries to find the classifier $f$ and encoding $\phi$ that minimizes this adversarial loss for the observed data. Our algorithm is adapted from \cite{madras2018learning}, where it was used for the purposes of \emph{fair representation learning}.

 To study the performance of the proposed approach on a newly coming domain $\cD_u \in \sD$, it is crucial to understand the behavior of our adversarial loss in the limit of large $k$ and $n_i$'s. However, the structure of the  discriminator changes with growing $k$. Thus, a crucial challenge lies in  pinning down whether our loss admits a limit, and if so, what should be the limiting value? Additionally, even if we can characterize this limit, how would the proposed algorithm perform on an arbitrary $D_u \in \mathcal{D}$? This paper explores these key questions in detail.

\paragraph{Contributions.}
 We obtain a precise characterization of the limit of our adversarial loss (Section \ref{sec:learning}). 
 We address the challenges incurred by the dependence of the discriminator on $k$ via a highly non-trivial geometric argument.  We then provide non-asymptotic generalization error bounds for the empirical loss around its population counterpart; the form of the population version is naturally determined using the prior limiting result. 
We further establish consistency of loss function optimizers $\hat{f}_{\lambda}, \hat{\phi}_\lambda$, in the
sense that, these converge (under an appropriate limit)  to the corresponding optimizers of the population loss.
Section \ref{sec:generalize} provides a characterization of the prediction performance of our algorithm on unseen domains that lie within bounded $\mathcal{H}$-divergence~\cite{Ben-David2010} of the seen ones. 
Section \ref{sec:invariant} decomposes 
our mappings $\phi$ into two components, and provides a complete characterization of {\em invariant} mappings (which defeat the discriminator) in terms of this decomposition.
Extensive experimental results are summarized in Section~\ref{sec:experiments}. 

\paragraph{Related Work.}

There are rich literatures of related work in computational learning theory.
For lack of space we confine our discussion to a handful of works in domain generalization.
In the earliest, kernel-based works on domain generalization~\cite{blanchard2011generalizing,muandet2013domain}, the learned classifier $f$ receives at test time not just a single $x$ drawn from a test distribution $D_T$, but (especially in~\cite{blanchard2011generalizing}) a large, unlabeled sample from $D_T$ together with a single additional test sample to be classified.  
To our knowledge, \cite{muandet2013domain} is the first to assume a latent distribution on domains (as do we). 

Three works are particularly aligned with our philosophical approach.  \cite{arjovsky2019invariant} comes from a line of work, initiated in~\cite{peters2016causal}, on causal inference and predictive robustness, relying on a notion of probabilistic invariance. (See~\cite{buhlmann2018invariance} for a survey.) 
\cite{arjovsky2019invariant} seeks 
data representations that elicit predictors satisfying certain invariance properties across the domains. This is framed as a penalized risk minimization problem, which is then solved using stochastic gradient descent. The theoretical guarantees rely on linearity assumptions \cite[Theorem 8]{arjovsky2019invariant}.

Inspired by~\cite{ganinetal}, adversarial networks were introduced for fair representation learning in~\cite{Edwards:2015vn,madras2018learning} and for domain generalization in~\cite{li2018domain,li2018deep}.
%
%
\cite{li2018domain} 
uses an autoencoder and introduces a Laplace prior on representations to encourage domain generalization. 
\cite{li2018deep} employs an adversarial architecture very similar to ours, expanded with a subnetwork
that seeks to minimize the discrepancy between $\mathbb{P}(X|Y)$ across the different domains,
addressing differences in base rates among the training distributions.
We provide theoretical insights not featured in~\cite{li2018domain,li2018deep}.
 



\section{Formal setup}\label{sec:setup}
\newcommand{\iid}{\stackrel{\mathrm{iid}}{\sim}}

Recall our setting from Section \ref{sec:introduction}. Throughout, we
assume that $\sD$ contains finitely many  probability distributions, i.e.  $\sD=\{\cD^*_1,\cD^*_2,\cdots,\cD^*_N\}$, and $\sD^{\cX}$ is the corresponding set of marginal distributions induced on $\cX$. Define $\cD_{1:k}$ to be the set of seen domains   $\{\cD_1,\hdots, \cD_k \}$, and assign them distinct ID's $\{1,\hdots,k  \}$. Let $S_{1:k} $ denote the collection of observed samples $\{ S_1,\hdots, S_k\}$. Note that   repeated sampling is possible here; for instance, we may have $\cD_1=\cD_2=\cD^*_1$.
Define $S^\cX_i:=\{x_{i,j}\}_{j=1}^{n_{i}}$ and $g(S^\cX_i):=\{g(x_{i,j})\}_{j=1}^{n_{i}}$, for any function $g$ on $\cX$.  For any function $g$ and distribution $\cD$, we use $g(\cD)$ to denote the distribution of $g(z)$, where $z\sim\cD$. 
For any distribution $\cD$ that admits a density function $p_{\cD}$, let $\mathrm{Supp}_{\cD}:= \{x| p_{\cD}(x) > 0 \}$.
Finally, for any function $f: \mathbb{R}^m \rightarrow \mathbb{R}^n$, we represent $f(\cdot)$ using the $n$-dimensional vector $(f^{(1)}(\cdot),\hdots, f^{(n)}(\cdot))^{\top}$. 

 \paragraph{Algorithm.} The samples $S_{1:k}$ are first passed through an encoder that produces a representation $\{\phi(S^{\cX}_i)\}_{i=1}^k$ of the input covariates. Here, $\phi$ is a representation mapping that belongs to some function class $\Phi= \{g | g: \mathbb{R}^d \rightarrow \mathbb{R}^s \}$. The output from the encoder is subsequently passed through a discriminator $\psi_k$ of the form 
\begin{equation}\label{eq:adversary}
 \psi_k(\cdot)=W\zeta(\cdot) +B,   
\end{equation}
where $\zeta : \mathbb{R}^s \rightarrow \mathbb{R}^p $ lies in some function class  $\Upsilon$, $W\in\bR^{k\times p}$ and $B\in\bR^k$. We further denote
$W=(w_1,w_2,\cdots,w_{k})^\top, B=(b_1,\cdots,b_k)^\top$,
where $w_i\in\bR^{p}, b_i\in \bR$.  Thus, the discriminator comprises a base structure $\zeta$ followed by a linear transformation, and effectively maps each input in $\mathbb{R}^s$ to $k$ unnormalized weights. For each input $\phi(x_{i,j})$, the $\ell$-th entry in the normalized version of the output $\psi_k(\phi(x_{i,j})) \in \mathbb{R}^k$ should be viewed as the discriminator's estimate of the probability that the pre-image $\phi^{-1}(\phi(x_{i,j}))$ was drawn from the seen domain with ID $\ell$. Finally, define $\pi_k(\cdot)$ to be the operation that maps an input vector $w$ to the index of the entry with maximal weight. If multiple entries  achieve the maximal weight, $\pi_k$ chooses uniformly among the corresponding indices. Simultaneously, a predictor is trained on the encoded representations $\{\phi(S^{\cX}_i)\}_{i=1}^k$ and produces labels in the outcome space. Denote the predictor class by $\cF=\{f \, | \, f: \bR^s\mapsto \cY \}$. 

\paragraph{Loss function.} The encoder, discriminator and predictor will be simultaneously trained using a loss function that comprises two components: (a) the loss corresponding to the predictor  
$
L_{\text{pred}}(\cD_{1:k},f,\phi)  =(1/k)\sum_{i=1}^{k}\bP_{(x,y)\sim\cD_i}(f(\phi(x))\neq y)$, (b) the loss corresponding to the discriminator or adversary $
L_{\text{adv}}(\cD_{1:k},\phi,\psi_k)  =\sum_{i=1}^{k}\bP_{x\sim\cD^\cX_i}(\pi_k\circ\psi_k(\phi(x))=i).$ The form of these loss functions is inspired from \cite{madras2018learning}.
Define 
\begin{equation}\label{eq:first-level-pop}
L(\cD_{1:k},f,\phi,\psi_k;\lambda)=L_{\text{pred}}(\cD_{1:k},f,\phi)+ \lambda L_{\text{adv}}(\cD_{1:k},\phi,\psi_k),
\end{equation}
where $\lambda>0$ is a tuning parameter, and the corresponding empirical version 
\begin{equation}\label{eq:empirical_loss}
    L(S_{1:k},f,\phi,\psi_k;\lambda)=\frac{1}{k}\sum_{i=1}^{k}\frac{1}{n_i}\sum_{j=1}^{n_i}\mI\{f(\phi(x_{i,j}))\neq y_{i,j}\}+\lambda\sum_{i=1}^{k}\frac{1}{n_i}\sum_{j=1}^{n_i} \mI\{\pi_k\circ\psi_k(\phi(x_{i,j}))=i\},
    \end{equation}
where $\mI$ is the indicator function. We seek to optimize the aforementioned loss to obtain
\begin{equation}\label{eq:optimization}
(\hat{f}_\lambda,\hat{\phi}_\lambda)=\arg \inf_{f\in\cF,\phi\in\Phi}\sup_{\psi_k\in\Psi_k}L(S_{1:k},f,\phi,\psi_k;\lambda).
\end{equation}
The infimum aims to maximize accuracy of the predictor, whereas the supremum ensures the performance of the discriminator is minimized. The final predictor for any test datapoint $x \sim \cD$ where $\cD \sim \mu$, is then given by 
 $\hat{y} := \hat{f}_{\lambda}(\hat{\phi}_\lambda (x))$. 
 

\begin{remark}
Recall the definition of $\cH$-Divergence \cite{Ben-David2010}: let $\cH$ be a class of binary classifiers, then $\cH$-divergence between distributions $\cD$ and $\cD'$ over $\bR^d$ is defined as 
$$D_\cH(\cD,\cD')=\sup_{h\in \cH}|\bP_{x\sim\cD}(h(x)=1)-\bP_{x\sim\cD'}(h(x)=1)|.$$
In the case of $k=2$, if we choose $\cH=\{\pi_2\circ \psi_2(\phi(\cdot)):\psi_2\in\Psi_2,\phi\in\Phi\}$, then we can see that 
$$\sum_{i=1}^{2}\bP_{x\sim\cD^\cX_i}(\pi_2\circ\psi_2(\phi(x))=i)=1+\bP_{x\sim\cD^\cX_1}(\pi_2\circ\psi_2(\phi(x))=1)-\bP_{x\sim\cD^\cX_2}(\pi_2\circ\psi_2(\phi(x))=1).$$
As a result, 
$$\sup_{\psi_2\in\Psi_2}\sum_{i=1}^{2}\bP_{x\sim\cD^\cX_i}(\pi_2\circ\psi_2(\phi(x))=i)=1+d_{\cH}(\phi(\cD^{\cX}_1),\phi(\cD^{\cX}_2)).$$
Our loss is a natural generalization of $\cH$- divergence and has a straightforward interpretation -- how best can the discriminator distinguish the images from the $k$ domains. We elucidate this further in Section \ref{sec:learning}.
\end{remark}

For simplicity, throughout the paper,  we only consider function classes for which the infimum and supremum can be achieved, and therefore replace $\inf$, $\sup$ in \eqref{eq:optimization} by $\min$, $\max$ respectively. In our experiments, the function classes $\cF, \Phi, \Psi_k$ are taken to be neural networks with specific architectures.

 \section{Theoretical Results}
\subsection{Learning theoretic analysis}\label{sec:learning}
To obtain a learning theoretic analysis of $L(S_{1:k},f,\phi,\psi_k;\lambda)$, it is crucial to understand its limiting behavior when the sample sizes $n_i$ and number of seen domains $k$ diverge. It is clear that when every $n_i \rightarrow \infty$, $L(S_{1:k},f,\phi,\psi_k;\lambda) \rightarrow L(\cD_{1:k},f,\phi,\psi_k;\lambda)$. Furthermore, $L_{\text{pred}}(\cD_{1:k},f,\phi) \rightarrow \bE_{
\cD\sim \mu}[\bP_{(x,y)\sim\cD}(f(\phi(x))\neq y)]$ as $k \rightarrow \infty$. Thus, we focus on studying the limiting behavior of $\max_{\psi_k\in\Psi_k}L_{\text{adv}}(\cD_{1:k},\phi,\psi_k)$ when $k$ diverges. Denote the density function of $\zeta(\phi(\cD^{*\cX}_i))$ by $\rho^{\zeta,\phi}_i(\cdot)$.


\begin{assumption}[Continuity] \label{assumption:nds}
Every 
$\zeta\in\Upsilon$ and $\phi\in\Phi$ is continuous almost everywhere (a.e.). Besides, for  all $
\cD^{*\cX}_i\in\sD$,$\zeta\in\Upsilon$ and $\phi\in\Phi$, $\rho^{\zeta,\phi}_i(\cdot) $ is continuous a.e. and $\mathrm{Supp}_{\zeta(\phi(\cD^{*\cX}_i))}$
 has non-zero volume in $\mathbb{R}^p$. 
\end{assumption}
Notice fully connected neural networks with ReLU activation functions are continuous a.e., since the ReLU activation function is  discontinuous only at $0$.
\begin{theorem}\label{thm:limit}
Suppose $\mu$ is a probability distribution supported on  a finite set of distributions $\mathcal{D}:=\{\cD^*_1,\cD^*_2,\cdots,\cD^*_N\}$, each of which is a distribution over $\mathcal{X} \times \mathcal{Y}$. Further, let $\cD_1,\cdots,\cD_{k} \iid \mu $ with corresponding study IDs $\{1, \hdots, k \}$.
Then under Assumption \ref{assumption:nds}, for any $\phi\in\Phi$,
\begin{equation}\label{eq:limit}
\lim_{k\rightarrow \infty}\max_{\psi_k\in\Psi_k}\sum_{i=1}^{k}\bP_{x\sim\cD^\cX_i}(\pi_k\circ\psi_k(\phi(x))=i)=\sup_{\cup_iA_i=\bR^p,A_i\cap A_j=\emptyset, \zeta\in\Upsilon}\sum_{i=1}^{N}\bP_{x\sim\cD^{*\cX}_i}(\zeta(\phi(x))\in A_i).
\end{equation}
\end{theorem}
The probabilities on the LHS (left hand side) are taken w.r.t.~marginal covariate distributions of the seen domains $\cD^{\cX}_i$, which may contain repeats from $\sD$. But the RHS contains probabilities w.r.t.~the corresponding marginals of all distributions in ${\sD}$. Speaking intuitively, Theorem \ref{thm:limit} says that, for {\em every} $\phi \in \Phi$, as $k$ grows so that (1) the encodings $\phi(\cD_{i}^{\mathcal{X}})$ contain repeated instances of every element from $\sD$, and (2) the structure of the last layer of the discriminator changes with $k$, the chance that the adversary accurately guesses the IDs of the encoded inputs is the same as the chance that the encoding $\phi(\cdot)$ itself maps the true distributions ${\cD_{i}^{\star}}^{\mathcal{X}}$ to $N$ disjoint parts of the space.

Theorem \ref{thm:limit} provides further insights into our loss function \eqref{eq:first-level-pop} and the behavior of our algorithm. Since the result holds for any $\phi \in \Phi$, when $k$ is large our algorithm effectively finds  $$(\hat{f}_\lambda,\hat{\phi}_\lambda) \approx \arg \min_{f,\phi} \{L_{\text{pred}}(\cD_{1:k},f,\phi)+ \lambda \sup_{\cup_iA_i=\bR^p,A_i\cap A_j=\emptyset, \zeta\in\Upsilon}\sum_{i=1}^{N}\bP_{x\sim\cD^{*\cX}_i}(\zeta(\phi(x))\in A_i)\}.$$ The second term is minimized for an encoding $\phi(\cdot)$ that maps the distributions ${\cD_i^{\star}}^{\mathcal{X}}$ to similar images, so that the adversary finds it difficult to guess the true IDs of the input covariates. (The prediction part of the loss discourages the trivial mapping $\forall x, \,\, \phi(x) = z$ for some arbitrary $z$.)

 The limit in \eqref{eq:limit} may be viewed as a measure of dissimilarity of the {\em set} $\{\phi({\cD_{i}^{\star}}^{\mathcal{X}})\}_{i=1}^N$.
In fact, consider  a setting where the supremum over $\zeta \in \Upsilon$ in the RHS of \eqref{eq:limit} is achieved and that the maximizer,  
$$\zeta^{\star}=\argmax_{\zeta\in\Upsilon}\sum_{i=1}^{N}\bP_{x\sim\cD^{*\cX}_i}(\zeta(\phi(x))\in A_i)$$
is unique. Then, it is not hard to see that 
$$\sup_{\cup_iA_i=\bR^p,A_i\cap A_j=\emptyset}\sum_{i=1}^{N}\bP_{x\sim\cD^{*\cX}_i}(\zeta^{\star}(\phi(x))\in A_i)\geq 1,$$ where equality holds iff $\zeta^{\star}(\phi(\cD^{*\cX}_1))=\cdots=\zeta^{\star}(\phi(\cD^{*\cX}_N)).$

If $\Upsilon$ contains the identity mapping, then  $$\sup_{\cup_iA_i=\bR^p,A_i\cap A_j=\emptyset}\sum_{i=1}^{N}\bP_{x\sim\cD^{*\cX}_i}(\zeta^{\star}(\phi(x))\in A_i) = 1$$ iff  $\phi(\cD^{*\cX}_1)=\cdots=\phi(\cD^{*\cX}_N).$  That is, the limit in \eqref{eq:limit} is minimized iff $\{\phi({\cD_{i}^{\star}}^{\mathcal{X}})\}_{i=1}^N$ are identical. 


To the best of our knowledge, such a precise understanding of the adversarial loss, as illuminated by Theorem \ref{thm:limit}, has so far eluded prior literature, and may be of independent interest for invariant representation learning \cite{wang2018invariant}. The fact that the structure of $\psi_k$ changes with $k$ presents significant challenges for the proof. We address this issue with a highly non-trivial geometric argument (Section A, Supplementary).
Speaking informally, we  cover the space $\mathbb{R}^p$ by grid cells such that, as $k$ increases, the number of cells grows and each cell becomes increasingly refined. As the cells grow finer, each one can be associated with an element from $\sD$ according to the distribution among $\{\zeta(\phi({\cD_{i}^{\star}}^{\mathcal{X}}))\}_{i=1}^N$ whose density in the cell is largest (ignoring ties). Then the final layer can be chosen such that, for every cell, the adversary assigns the highest weight to the  corresponding distribution.

\begin{remark}
For $N=2$, the limit can be related to the total variation distance since $$\sup_{\cup_iA_i=\bR^p,A_i\cap A_j=\emptyset}\sum_{i=1}^{2}\bP_{x\sim\cD^{*\cX}_i}({\zeta}^*(\phi(x))\in A_i)=TV({\zeta}^*(\phi(\cD^{*\cX}_1)),{\zeta}^*(\phi(\cD^{*\cX}_2)))+1.$$
\end{remark}

\paragraph{Non-asymptotic generalization bounds.} 
We now turn to bounding the generalization error of $L(S_{1:k},f,\phi,\psi_k;\lambda)$. To this end, a few key quantities are introduced next. Define
\begin{align}\label{eq:population}
L(\sD,f,\phi;\lambda) \nonumber&=\bE_{
\cD\sim \mu}[\bP_{(x,y)\sim\cD}(f(\phi(x))\neq y)]\\
&+ \sup_{\cup_i A_i=\bR^p,A_i\cap A_j=\emptyset, \zeta\in\Upsilon}\lambda\sum_{i=1}^{N}\bP_{x\sim\cD^{*\cX}_i}(\zeta(\phi(x))\in A_i).
\end{align}
\begin{definition}[Grid Cells]\label{def:grid}
For $n \in \mathbb{N}^{+}, B \in \mathbb{R}^{+}$, define  $G(n,B)$ to be the set
$$G(n,B)=\{I_{i_1}\times I_{i_2}\times\cdots\times I_{i_p},i_j\in\{1,\cdots,n\} \,\, \forall \,\,j\},$$
where $I_{i_j}=[-B+2(i_j-1)B/n,-B+2i_jB/n]$.
\end{definition}

The elements in $G(n,B)$ form a partition of  $[-B,B]^{p}$ and the intersection of every pair of elements has volume $0$ in $\mathbb{R}^p$. Now, let $H_k$ be a collection of distributions in $\sD$ that receive  high $\mu$-probability in the following sense, $H_k := \{\cD_i \in \sD : \mu(\cD_i) \geq 1/k^{1/4} \}$.  Define $T_i^{\zeta,\phi}$ to be the set of points in $\mathbb{R}^p$ where the density $\rho_i^{\zeta, \phi}$ is maximized (up to ties), that is,
$$T^{\zeta,\phi}_i:=\{z \in \mathbb{R}^p:\rho^{\zeta,\phi}_i(z)>\rho^{\zeta,\phi}_j(z),~\text{for all}~j< i~\text{and}~\rho^{\zeta,\phi}_i(z)\geq\rho^{\zeta,\phi}_j(z),~\text{for}~j\geq i\}.$$
It is easy to see that $\{T^{\zeta,\phi}_i\}_{i=1}^{N}$ form a partition of $\bR^p$. Furthermore, let $M^{\zeta,\phi}_{i,1}(n,B)$ denote the collection of grid cells on the boundary of $T_i^{\zeta,\phi}$, and $M^{\zeta,\phi}_{i,2}(n,B)$ denote those in the interior. 
\begin{align*}
   M^{\zeta,\phi}_{i,1}(n,B)  &=\{g\in G(n,B)| g\nsubseteq T^{\zeta,\phi}_i,g\cap T^{\zeta,\phi}_i\neq\emptyset\}\\
   M^{\zeta,\phi}_{i,2}(n,B) & =\{g\in G(n,B)| g\subseteq T^{\zeta,\phi}_i\}. 
\end{align*} 
\begin{assumption}
[Boundedness]
\label{assumption:ub}
There exists a constant $B_{\rho} $ and function $B(\cdot)$ s.t.~for any $\varepsilon>0$, 
$\sup_{\zeta,\phi}\sum_{i=1}^{N}\bP_{x\sim \cD^{*\cX}_i}(\|\zeta(\phi(x))\|_2\geq B(\varepsilon))\leq \varepsilon$ and  $\sup_{z,\zeta,\phi,i}|\rho^{\zeta,\phi}_i(z)|\leq B_\rho.$

\end{assumption}
\begin{assumption}[Bounded VC-dimensions]\label{assumption:vc}
Assume that the function classes $\Lambda=\{f\circ \phi|f\in\cF,\phi\in\Phi\}$  and $\Xi = \{\mI\{w_1^\top\zeta( \phi(x))+b_1>w_2^\top\zeta( \phi(x))+b_2\}| w_i\in \bR^p, b_i\in \bR,\zeta\in\Upsilon,\phi\in\Phi,i=1,2\}$ have VC-dimensions $\cV_{\Lambda}$ and $\cV_{\Xi}$ respectively. 
\end{assumption}

Note that in Assumption \ref{assumption:vc}, the VC dimension condition on $\Xi$ is on two nodes instead of $k$.


\begin{theorem}\label{thm:non-asymptotic bound}
Consider the setting of Theorem \ref{thm:limit}, and define $m_k :=\lceil k^{\frac{3}{4}}-\sqrt{(k\log(|H_k|)+k^{\frac{3}{4}})/ \sqrt{2}}\rceil$. 
Under Assumptions \ref{assumption:nds}-\ref{assumption:vc}, there exists a universal constant $c$, s.t.~for any $t_1,t_2>0$, w.p.~at least  $1-e^{-k^{1/4}}-\sum_{i=0}^{k-1}4e^{-n_it_1^2}-2Ne^{-2kt_2^2}$, 
\begin{align}\label{eq:na}
\begin{split}
& \max_{f\in\cF,\phi\in\Phi}|\max_{\psi_k\in\Psi_k} L(S_{1:k},f,\phi,\psi_k;\lambda)-L(\sD,f,\phi;\lambda)|
\leq (1+k\lambda)t_1+\frac{2\lambda}{\sqrt{k}}+N\cdot t_2+ \text{I}+  \text{II}+ \text{III},\\
\end{split}
\end{align}
where $\cV_{\cC(k)}=k\cV_{\Xi}(\log(\cV_{\Xi}))^2$, and
\begin{align*}
\text{I}=\lambda\max\{N &-|H_k| ,0\} ,\quad\text{II}=\frac{2\lambda B_\rho \left(B(\frac{1}{\sqrt{k}})\right)^{p}}{\lfloor m^{1/p}_k \rfloor^p } \sum_{i\in H_k}\sup_{\zeta,\phi}|M^{\zeta,\phi}_{i,1}(\lfloor m^{1/p}_k \rfloor,B(\frac{1}{\sqrt{k}}))|,\\
&\text{III}=\frac{2c}{k}\sum_{i=1}^{k}\frac{k\sqrt{\cV_{\cC(k)}\log(\frac{n_i}{\cV_{\cC(k)}})}+\sqrt{\cV_{\Lambda}\log(\frac{n_i}{\cV_{\Lambda}})}}{\sqrt{n_i}}.
\end{align*}
\end{theorem}

 We now proceed to analyze the bound in \eqref{eq:na}. Note that III vanishes when $\min_{i} n_i =\Omega( k^\alpha)$ for $\alpha\geq 2$, whereas $\text{I}$ is small when $k$ is much larger than $N$. 
 
For II, note that for $k$ much larger than $N$, $\log(|H_k|)$ is negligible compared to $\sqrt{k}$, so that $m_k =  \Omega (k^{3/4})$. Now for fixed $B$, when $k$ is large, $G(\lfloor m^{1/p}_k \rfloor,B)$ shrinks in volume.  In settings where the union of the grid cells in $ M^{\zeta,\phi}_{i,2}(\lfloor m^{1/p}_k \rfloor,B)$ approximates $T_i^{\zeta,\phi}$ well enough with growing $k$, $M_{i,1}^{\zeta,\phi}(\lfloor m^{1/p}_k \rfloor,B)$ contains negligible number of grid cells compared to $ M^{\zeta,\phi}_{i,2}(\lfloor m^{1/p}_k \rfloor,B)$, leading to $$\sum_{i\in H_k}\sup_{\zeta,\phi}|M^{\zeta,\phi}_{i,1}(\lfloor m^{1/p}_k \rfloor,B)|=o(\lfloor m^{1/p}_k \rfloor^p).$$We defer the readers to the Supplementary Section A for specific examples demonstrating this phenomenon. This continues to hold when $B$ is replaced by $B(1/\sqrt{k})$ if the latter grows slowly with increasing $k$. Since $B(\cdot)$ is related to the tails of the distributions $\zeta(\phi(\cD^{*\cX}_i))$, we are able to control this term in specific examples. For instance, if all distributions in $\sD$ are sub-Gaussian with sub-gaussian norm bounded by some constant $\sigma_{\max}$, then $B(1/\sqrt{k})=O(\sqrt{\log k})$. Together, this means that II is also small when $k$ is sufficiently large. Thus, Theorem \ref{thm:non-asymptotic bound} demonstrates that observing samples from \emph{more domains helps in generalization}. 

\paragraph{Consistency.}
 Theorem \ref{thm:non-asymptotic bound} provides conditions on $k$ and $n_i$, $i \in [k]$, under which the empirical loss function, when evaluated at the estimates $(\hat{f}_\lambda, \hat{\phi}_\lambda)$, will be close to its population counterpart w.h.p. Here we seek to  establish that these estimates, in fact, well approximate the minimizers of the population loss. Since we impose no assumptions on the specific distributional forms of the seen domains, this is hard to prove in such generality. We will therefore establish this under a curvature condition on the population loss that is slightly weaker than strong convexity.

\begin{assumption}[Well-separation]\label{assumption:ws}
Denote $\cM^*_{\cF,\Phi}\subseteq\cF\times\Phi$ to be the set of minimizers of $L(\sD,f,\phi;\lambda)$. For a metric $dist(\cdot,\cdot)$ on the function class $\cF\times\Phi$, there exists a function $U(\cdot;\lambda):\bR\rightarrow\bR^{+}$ satisfying $\lim_{\varepsilon\rightarrow 0}U(\varepsilon;\lambda)\rightarrow 0$, such that for any $\varepsilon>0$
$$\inf_{\xi \in\cF\times\Phi: ~\inf_{z\in\cM^*_{\cF,\Phi}}dist(\xi,z)\geq U(\varepsilon;\lambda)}|L(\sD,\xi;\lambda)-\min_{f\in\cF,\phi\in\Phi}L(\sD,f,\phi;\lambda)|\geq \varepsilon.$$
\end{assumption}
\begin{theorem}\label{thm:minimizer}
Under Assumption \ref{assumption:nds}-\ref{assumption:ws}, almost surely,
$$\inf_{z\in\cM^*_{\cF,\Phi}}dist((\hat{f}_\lambda,\hat{\phi}_\lambda),z)\leq U(2\Gamma;\lambda),$$
where $\Gamma$ equals the RHS of \eqref{eq:na}.
\end{theorem}




\subsection{Generalization to unseen domains}\label{sec:generalize}
Theorem \ref{thm:minimizer} establishes that, under the aforementioned conditions, our proposed classifier $\hat{f}_\lambda(\hat{\phi}_{\lambda}(\cdot))$  minimizes the population loss $L(\sD,f,\phi;\lambda)$. However, this loss 
is a penalized version of the expected prediction error under $\mu$.  Naturally, the results from the preceding section fail to capture the behavior of our classifier on an arbitrary domain from $\sD$. We now address this problem, showing that such a worst case characterization is possible if elements in $\sD$ are well-represented under $\mu$---that is, every domain in $\sD$ is close to at least one domain that receives relatively high $\mu$-probability.


\begin{assumption}[Well-represented]\label{assumption:rep}
There exists constants $0<p_l<1$ and $\delta>0$, s.t.~for any $\cD\in\sD$ with $\mu(\cD) > 0$, $\exists \cD'\in\sD$ with $\mu(\cD') \geq p_l$ and
$d_\cH(\cD,\cD')\leq\delta,$
where $\cH=\cF\times\Phi$.
\end{assumption}

\begin{theorem}\label{thm:worstcase}
Under Assumptions \ref{assumption:nds}, \ref{assumption:vc} and \ref{assumption:rep}, w.p.~at least $1-\exp(-k^2p_l^2/2)/p_l-\sum_{i=1}^{k}4e^{-n_it^2}$ over the randomness in $S_{1:k}$ and $\cD_{1:k}$, for any $\cD_u\in\sD$ and all $f\in\cF,\phi\in\Phi$,
$$\bP_{(x,y)\sim\cD_u}(f(\phi(x))\neq y)\leq\frac{2}{p_l} \Big(\hat{\beta}(f,\phi)+t+c\sqrt{\frac{\cV_{\Lambda}\log(n_i/\cV_{\Lambda})}{n_i}}\Big)+\delta,$$
where $\hat{\beta}(f,\phi)=\sum_{i=1}^{k}\sum_{j=1}^{n_i}\bI\{f(\phi(x_{i,j}))\neq y_{i,j}\}/(kn_i).$ Moreover, this holds even if  $|\sD|$ is countably infinite. 
\end{theorem}




\subsection{Characterization of invariant representation mappings}\label{sec:invariant}
\begin{definition} \label{def:invariant}
An element $\phi \in \Phi$ is said to be an invariant representation mapping for a collection of $k$ domains $\tilde{\cD}_1,\hdots, \tilde{\cD}_k \in \sD$ and for some $\epsilon > 0$, if 
 $\sup_{\psi_k\in\Psi_k}\sum_{i=1}^k\bP_{x\sim\tilde{\cD}_i^{\mathcal{X}}}(\pi_k\circ\psi_k(\phi(x))=i)\leq \varepsilon$. 
\end{definition}
Recall that the range of any $\phi\in\Phi$ is $\mathbb{R}^s$ so that $\phi(\cdot)$ may be expressed in the form  $(\phi^{(1)}(\cdot),\phi^{(2)}(\cdot),\cdots, \phi^{(s)}(\cdot))^\top$. Suppose that the space containing $\phi^{(i)}(\cdot)$ is \textbf {separable}, that is, there exists basis functions $\{\beta_j\}_{j=1}^m$ ($m$ can be infinity), such that $\forall i\in\{1,\cdots,s\}$, 
$$\phi^{(i)}(x)=\sum_{j=1}^m \alpha^{(i)}_{j}(\phi)\beta_j(x).$$
On defining the matrix $M_\phi=\{\alpha^{(i)}_{j}(\phi)\}_{ij}$, we have 
$$\phi(x)=M_\phi\cdot(\beta_1(x),\beta_2(x),\cdots, \beta_m(x))^\top$$ and let  
$$\Gamma(x)=(\beta_1(x),\beta_2(x),\cdots, \beta_m(x))^\top.$$
Denote $M_{\phi}^{-}$ to be the MP-inverse of $M_{\phi}$. Finally, for any $\psi_k \in \Psi_k$, define $$I_i(\psi_k)=\{z:\psi^{(i)}_{k}(z)>\max_{j\neq i}\psi^{(j)}_k(z)\}.$$

With this decomposition we can now characterize invariant mappings.

\begin{theorem}\label{thm:existencemultiple}
For any $\varepsilon>0$, if $\forall \psi_k \in \Psi_k$, 
$\cup_i I_i(\psi_k) = \bR^s$, then $\phi \in \Phi$ satisfies Definition \ref{def:invariant} iff 
\begin{align*} 
 \exists f \in \text{Ker}(M_\phi) \,\, \text{s.t.~} \,\, \sum_{i=1}^k\bP_{x\sim\tilde{\cD}_i}(\Gamma(x)+f(x)\in M^{-}_\phi I_i(\psi_k))\leq\varepsilon.
 \end{align*}
\end{theorem}
Above, the condition $\cup_i I_i(\psi_k) = \bR^s$ is necessary to ensure that there will be no ties between the $k$ weights produced by $\psi_k$. Note that our previous results do not require this condition. 



\section{Experiments}\label{sec:experiments}
\label{sec:experiments}
We assessed the performance of our  approach on several datasets: (a) synthetic data based on those in biomedical studies  \cite{Patil:2018gr}, (b) colored MNIST  \cite{Deng:mnist},  (c) PACS  \cite{li2017deeper}. Our experiments confirm the conclusion (Section \ref{sec:learning}) that observing more domains  improves generalization  performance on an unseen one. 
For (a), we compared with logistic regression and random forest,  whereas (b) and (c) were benchmarked against the state-of-the-art algorithms, IRM \cite{arjovsky2019invariant} and CIDDG \cite{li2018deep}. Our code was adapted from the LAFTR code base \cite{madras2018learning}, but the decoder was dropped to be constistent with our theoretical setting\footnote{Extending the theory to include the decoder is an interesting direction for future work.}. The prediction loss was taken to be binary cross-entropy.




\paragraph{Synthetic Data.}
We consider synthetic data settings with $k=4$ and $k=10$. In each case, to sample a data point from a domain $\cD_i$, a pair $(x_j,y_j)$, $x_j \in \mathbb{R}^{30}, y_j \in \{0,1 \}$ is generated with $x_j \sim \mathcal{N}(\mu_i, \Sigma_i)$. The outcome $y_j$ is generated so that a part of the relationship between $x_j$ and $y_j$ remains invariant across domains, while the other part varies. To operationalize this, we select a random subset $\mathcal{A}$ of covariates, a base rate $b_i$, and a set of functions $\{f_{\mathrm{inv}}, f_1,\hdots, f_k$\}. We then sample $y_j \sim \mathrm{Ber}(b_i)$ and accept $(x_j,y_j)$  if $y_j=\bm{1}(f_{\mathrm{inv}}(x_{j,\mathcal{A}}, \epsilon_{j,i}) > 0) = \bm{1}(f_i(x_{j,{\mathcal{A}^c}}) > 0) $, where $\epsilon_{j,i} \sim F_i$ is an additional small error term.  Here, the parameters $\mu_i, \Sigma_i, b_i,F_i, f_i$ vary between the domains whereas $f_{\mathrm{inv}}$ and $\mathcal{A}$ remain invariant;  $\epsilon_{j,i}$ ensures that the invariant signal between domains is not strong compared to the domain-specific one. Table \ref{table:synthetic_baseline} reports the performance of our algorithm on a new unseen domain of the same  form as above, but with different parameters $\mu_{k+1}, \Sigma_{k+1}, b_{k+1},F_{k+1}, f_{k+1}$. (Training involved $5000$ samples from each seen domain.) Observe that the test accuracies increase from $k=4$ to $k=10$. We uniformly outperform both baselines by a notable margin. 
\begin{table} [H]
 \caption{Test domain classification accuracy on  (a) synthetic data where each of the functions $f_{\mathrm{inv}}, f_1,\hdots, f_k, f_{k+1}$ contain a linear component and an interaction term; (b) similar synthetic data with responses generated differently (Section B, Supplementary) and each of the aforementioned functions now contain logical OR of two linear functions. }
 \label{table:synthetic_baseline}
 \centering
 \begin{tabular}{lllll}
    \toprule
    Algorithm    & (a) 4-Domain    & (a) 10-Domain & (b) 4-Domain  & (b) 10-Domain \\
    \midrule
    RVR & \textbf{90.6}\% & \textbf{95.6}\% & \textbf{86.1}\% & \textbf{93.4}\% \\
    Logistic Regression & 82.3\% & 86.2\% & 82.6\% & 86.7\% \\
    Random Forest & 79.4\% & 89.0\% & 85.0\% & 88.3\% \\
    \bottomrule
 \end{tabular}
\end{table}
\paragraph{Colored MNIST.}
The colored MNIST data was generated from the  MNIST database on handwritten digits \cite{Deng:mnist}; here, the digit color acts as a spurious signal and the digit shape acts as the invariant signal. We perform binary classification on several versions of this dataset, following experimental setups similar to \cite{arjovsky2019invariant}. In Table \ref{comparetable}, "\textbf{$A\%$-shape $B\%$-color}" refers to a setting with two training domains ($10,000$ samples each) both containing $A\%$ correlation between digit shape and  labels: digits $0-4$ receive label $0$ w.p.~$A\%$ ($1$ o.w.), and digits $5-9$ receive label $1$ w.p.~$A\%$ ($0$ o.w.).
In addition, there is a $B\%$ domain-specific correlation between digit color and labels: domain $1$ (resp. domain ~$2$) associates the color red with label $0$ (resp.~$1$) and green with label $1$ (resp.~$0$) w.p.~$B\%$. The last setting in Table \ref{comparetable} contains $6$ training domains with shape-label correlation similar to that for the first, but the color-label correlation varies largely across domains, with each one consisting of mixtures of 2-3 different digit colors. The unseen test domains constitute either single color digits or a random mixture of red-green digit colors assigned independent of the label, and the same shape-label correlation as the corresponding training data. Our algorithm beats IRM and CIDDG in multiple settings, and performs comparably in others. Finally, Table 2 reports our performance when only $3$ of the $6$ domains from this setting are used as the training data (same test data). Once again, test accuracy improves remarkably with more seen domains.

\begin{table}
\caption{Multi-domain comparison of test accuracy on colored MNIST with $100\%$ digit-label correlation and varying color-label correlations. The second column is the same as  Table \ref{comparetable}, row 3.}
  \label{colormulti}
  \centering
  \begin{tabular}{lll}
    \toprule
      & 3-Domain     & 6-Domain \\
    \midrule
    RVR & 86.1\% & 97.7\% \\ 
    \bottomrule
  \end{tabular}
\end{table}



\begin{table} [H]
  \caption{Test accuracy on several colored MNIST settings. \textit{Target} denotes the digit color of the test domain. Details of the color-label correlation for row 3 can be found in Section B, Supplementary. }
  \label{comparetable}
  \centering
  \begin{tabular}{lllllll} 
    \toprule
    & Setting & Target & $k$ & RVR & IRM & CIDDG\\
    \midrule
    1. & $100\%$-shape $90\%$-color & purple & 2 & \textbf{97.5}\% & 94.3\% & 95.7\% \\
    \midrule
    2. & $75\%$-shape $80\%$-color & red-green & 2 & \text{69.7*}\% & 69.0\% & \textbf{71.1}\% \\
    \midrule
    3. & $100\%$-shape, unequal color & white & 6 & \textbf{97.7}\% & 94.7\% & 96.9\% \\
    \bottomrule
  \end{tabular}
\end{table}

\paragraph{PACS.}
To conclude, we examine our  performance on PACS, an image-style dataset made of Photos, Art, Cartoon, and Sketch, which has been repeatedly used \cite{li2018domain}  to benchmark domain generalization algorithms. 
We specifically consider images labeled giraffes (label 0) or elephants (label 1), which leads to $384, 540, 803, 1493$ samples respectively. Each domain alternates as the target domain, while the algorithm trains on the rest.
Once again, our algorithm beats (Table \ref{pacstable}) both  baselines across the board, and by a significant margin in most settings.

\begin{table}[H]
    \caption{Test accuracy on two types of images obtained from PACS }
  \label{pacstable}
  \centering
  \begin{tabular}{llll}
    \toprule
    Target & RVR & IRM & CIDDG\\
    \midrule
    P & \textbf{70.7}\% & 57.6\% & 62.1\% \\
    \midrule
    A & \textbf{66.7}\% & 64.2\% & 59.9\% \\
    \midrule
    C & \textbf{80.8}\% & 75.0\% & 73.8\% \\
    \midrule
    S & \textbf{54.3}\% & 54.0\% & 53.4\% \\
    \bottomrule
  \end{tabular}
\end{table}



\section{Discussion}

One natural question is whether we can extend the theoretical results to cross-entropy or other notions of loss. Next, our theoretical analysis does not yet fully capture our intuition regarding the conditions under which we believe our approach will succeed. 

Our work forges a new path to address a major problem in biomedical research, where high-dimensional datasets are frequently encountered, and predictive algorithms  increasingly used to inform personalized medical care.  While strategies to ensure generalizability of these algorithms beyond the populations studied during training have been lacking, we are encouraged by our experimental results and have initiated engagement with the biomedical applications that inspired this work.

\section*{Acknowledgements}
This work was supported in part by the Center for Research on Computation and Society (Harvard SEAS), the Harvard Data Science Initiative, NSF CCF-1763665, NIH/NCI 5T32CA009337-39, and NSF-DMS 1810829.

\newpage






\bibliography{UD_cite,rvr,rvrps}
\bibliographystyle{plain}

\end{document}